\documentclass[article]{elsarticle}                           %
\usepackage{graphicx}
\usepackage{epsfig}
\usepackage{algorithm,algorithmic}
\usepackage{amsmath}
\usepackage{graphicx, array, blindtext}
\usepackage{enumitem}
\usepackage{booktabs}
\usepackage{multicol}
\usepackage{blindtext}
\usepackage{mathtools, cuted}
\usepackage{graphicx}
\usepackage{tabularx} 
\usepackage[usenames, dvipsnames]{color}
\usepackage{times}
\usepackage{soul}
\usepackage{url}
\usepackage[utf8]{inputenc}
\usepackage[small]{caption}
\usepackage{graphicx}
\usepackage{amsmath}
\usepackage{booktabs}
\begin{document}

\begin{frontmatter}
\title{Fire Threat Detection From Videos with Q-Rough Sets}

\author[mymainaddress]{Debarati B. Chakraborty\corref{mycorrespondingauthor}}
\cortext[mycorrespondingauthor]{Debarati B. Chakraborty}
\ead{debarati.earth@gmail.com}

\author[mymainaddress]{Vinay Detani}
\ead{detani.1@iitj.ac.in}

\author[mymainaddress]{Shah Parshv Jigneshkumar}
\ead{shah.7@iitj.ac.in}

\address[mymainaddress]{Dept. of Computer Science and Engineering, Indian Institute of Technology, Jodhpur, India}

\begin{abstract}
This article defines new methods for unsupervised fire region segmentation and fire threat detection from video stream. Fire in control serves a number of purposes to human civilization, but it could simultaneously be a threat once its spread becomes uncontrolled. There exists many methods on fire region segmentation and fire non-fire classification. But the approaches to determine the threat associated with fire is relatively scare, and no such unsupervised method has been formulated yet. Here we focus on developing an unsupervised method with which the threat of fire can be quantified and accordingly generate an alarm in automated surveillance systems in indoor as well as in outdoors. Fire region segmentation without any manual intervention/ labelled data set is a major challenge while formulating such a method. Here we have used rough approximations to approximate the fire region, and to manage the incompleteness of the knowledge base, due to absence of any prior information. Utility maximization of Q-learning has been used to minimize ambiguities in the rough approximations. The new set approximation method, thus developed here, is named as Q-rough set. It is used for fire region segmentation from video frames. The threat index of fire flame over the input video stream has been defined in sync with the relative growth in the fire segments on the recent frames. All theories and indices defined here have been experimentally validated with different types of fire videos, through demonstrations and comparisons, as superior to the state of the art.

\end{abstract}

\begin{keyword}
Fire segmentation\sep video processing \sep rough sets \sep Q learning \sep fire threat detection \sep granular computing
\end{keyword}
\end{frontmatter}

\section{Introduction}
Fire is both a threat and a necessity of mankind. We light fires to fulfill our necessity (e.g., for cooking, generating heat), simultaneously it could be a threat if it becomes uncontrolled. We know that there are some objects that are easily flammable (e.g., cloths, plastics) and some that burns out gradually (e.g., candle, wood or logs in the fireplace etc.). We burn the latter kind of elements to meet our need in a controlled environment, whereas fires concerning the former group of elements could be dangerous. Fire spreads quite fast in the flammable objects, and from one several other objects catch the fire. Therefore, measuring the spread of the fire could be a solution in automated fire detection systems to find out the possible threat of the visible fire. Here we tried to come up with a solution so that the fire threat can be detected automatically only with a surveillance camera, and no other sensors will be required. Besides, we tried to make the method fit for both indoor and outdoor surveillance. 

Proper classification of fire region(s) from the non-fire ones and automated segmentation of the fire region is a very important step while quantifying the threat of the fire. The aim of this work could be summarized as follows: i) to develop an unsupervised method for fire pixel classification, ii) to identify fire regions from different varieties of videos, and iii) to determine the threat associated with the fire flame over the input video stream. Here in this article we have formulated a new method of fire region segmentation in videos using hybridization of rough sets and Q-learning. Besides, we have also defined a new measure, namely, fire threat index, to quantify the threat associated with a fire from video streams. Before discussing the other methods on vision-based fire detection, here in this section we shall roughly discuss the key concepts of rough set theory and Q-learning relevant to this article.

 Theory of rough sets, as explained by Pawlak
\cite{PAWLAK_RS_BOOK_1992}, has become a popular mathematical
framework for granular computing. The focus of the theory is on the
ambiguity caused by limited discernibility of objects in the domain
of discourse. Its key concepts are those of object
'indiscernibility' and 'set approximation'. Two major
characteristics of the theory that have drawn the attention of
applied researchers are uncertainty handling (using lower and upper
approximations) and granular computing (using information granules). Theory of rough sets has been proven to be successful in many areas, like, feature extraction from video streams with information flow \cite{Pal_17}, on-line multi-label feature selection from streaming data \cite{Liu_18}, and information entropy based fast feature selection for big data analytics \cite{Zhaoa_20}.  We have aimed to develop an unsupervised fire detection method here in this article. Since we have limited access to the knowledge-base while identifying fire in a video frame, we have approximated the fire region with lower and upper approximations of rough sets.

Q-learning is a reinforcement learning algorithm \cite{Hasslet_10}. The primary difference between Q-learning and other reinforcement learning techniques is that it is model-free \cite{Sutton18}, i.e., here the agent does not need any state transition model either for learning or for action selection. Q-learning prefers the best Q-value from the state reached in the observed transition \cite{russel2020}. The actual policy, being followed by the agent, is not taken into account in Q-learning, that is why, it is also known as off-policy control. In Q-learning, the agent learns the action utility function (or Q-function) to estimate the expected utility of choosing a certain action when it is in a particular state.

Here in this work we are not considering any prior information on the fire region or possible spread of the fire pixels. All information is to be determined by the unsupervised process itself. Therefore, the proposed solution has an initial rough approximation of fire regions, and then employs a Q-agent to minimize the ambiguities that is present is the rough approximation, with utility maximization. This is the broad idea underlying Q-rough set with utility-based agent. States, in which the Q-agents are supposed to be in, are different granules (clump of data points \cite{Zadeh_97}) and actions are insertion or deletion of those granules from fire regions. The same Q-agents are then employed over the video stream to extract out the fire regions in each frame. The fire threat index is then formulated using the segmented fire regions over the video stream. 

The novelty of the technique described in the article can be summarized as: i) definition and application of the Q-rough set (that use unsupervised set estimations) to minimize ambiguities in rough approximations, ii) segmentation of fire regions from a video frame with Q-rough set, iii) employing Q-agents over a sequence of frames to extract fire regions in each frame, and iv) formulation of a fire threat index to quantify the threat of fire flame in a video stream. All the theoretical formulations have been experimentally verified and proven to be effective over the state of the art methods.

The rest of the article is organised as follows. A few state of the art methods for fire detection are discussed in Section \ref{RWF}. The underlying steps of the proposed work is described in Section \ref{PWF}. Q-rough set is defined in Section \ref{FQR} alongwith brief descriptions of rough sets and Q-learning. The method of fire region segmentation with Q-rough set is developed in Section \ref{QRSFD}. The quantification of the threat of fire flame is carried out in Section\ref{DTF}. The qualitative and quantitative experimental results of the proposed methods are given in Section \ref{res} alongwith suitable comparative studies. The overall conclusions of this article are drawn in Section \ref{conF}.  

\section{Related Work}\label{RWF}
The problem of fire detection from images is being addressed over a decade \cite{Gaur_20}. Here we are going to discuss a few bench-mark methods. The problem was first addressed by Chen \emph{et al.} \cite{Chen_04} where fire was detected with RGB values and rule-base. Ferneds \textit{et al.} then developed a method for forest fire detection by classifying lidar signals with neural network committee machine in \cite{Fernandesa_04}. Flame detection by modelling the RGB distributions of fire -pixels with mixture of Gaussian (MoG) model \cite{Toreyin_05}, and with motion-information and hidden Markov model (HMM) \cite{Toreyin_06} were then done by Toreyin \emph{et al.} Fire pixel identification in videos was then carried out by Celik \cite{Celik_10} where CIE L*a*b color space was used to identify fire pixels. The method of forest fire detection by incorporating static and dynamic features in videos in HSV color space was proposed by Zhao \emph{et al.} in \cite{Zhao_11}. Chino \emph{et al.} then developed a way of fire detection from still images by combining color and texture features in \cite{Chino-15}. The method of fire detection with spatio-temporal consistency energy of each candidate fire region was estimated by Dimitropoulos \emph{et al.} \cite{Dimitropoulos_15} with prior knowledge about the possible existence of fire in neighboring blocks from the current and previous video frames, and an SVM-based classification of fire-non fire regions was also executed there.  Recently a couple of fire detection method has been developed with deep learning techniques \cite{Li_20}. Muhammad \emph{et al.} \cite{Muhammad_18} came up with a solution by incorporationg deep features of CNNs in fire-detection and high-priority cameras based on cognitive radio networks were developed. Kim and Lee \cite{Kim_19} developed a method of detecting fire with faster region grow method using deep CNN (convolutional neural network). Cai \textit{et al.} \cite{Cai_19} formulated an improved deep CNN that uses the global average pooling layer instead of the full connected layer to fuse the acquired depth features and detect fire.  

All the methods, we have discussed so far, either requires some initial information about fire pattern or needs initial manual labelling for training. Most of the methods are focused on some specific applications, such as, forest fire, indoor fire, outdoor fire, etc. The method that we developed here is completely unsupervised and does not need any prior information/ manual intervention. Besides, it is very effective in classifying fire pixels in any types of video frame, that is it's a general method for fire segmentation, which could be applicable to anywhere for fire detection. The method of quantification of fire threat is also new to literature. We are going to describe the basic steps of the method in the following section.

\section{Proposed Work}\label{PWF}
\begin{figure}
    \centering
    \includegraphics[width=5in]{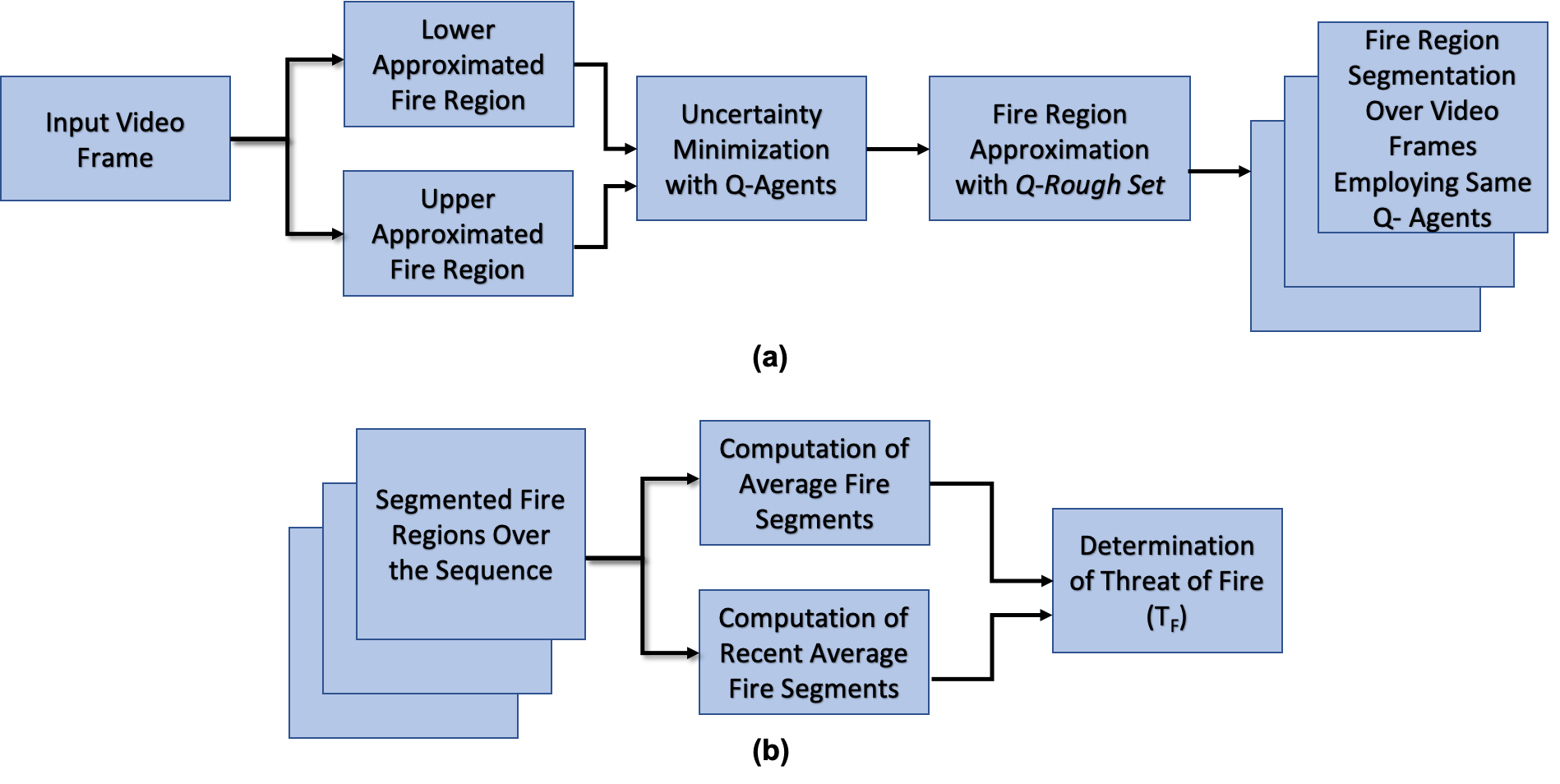}
    \caption{Block diagram of (a) fire region segmentation and (b) threat detection methods }
    \label{BlockDia}
\end{figure}

Our work can be subdivided into two parts, viz. i) detection of fire from videos and ii) determination of spreading of fire or quantifying the threat associated with the fire. The step-wise formulations of the two methods are shown in Figs. \ref{BlockDia}(a) and \ref{BlockDia}(b).  

In Fig. \ref{BlockDia}(a) proposed fire-flame detection method is shown, that will accumulate information from different color space and collate the information judiciously to classify the fire pixels correctly from the video frames.  Here Q-learning is rough because the state-action policy is determined with the lower approximated regions (obvious fire region here), and the ambiguities between the two approximations get minimized by utility maximization of the Q-agents. The same Q-agents are then employed over the rest of the video sequence to have faster decision making over known states. This part is described in details in Section \ref{QRSFD}. 

In the second part of the work we aim to quantify the threat associated with the fire flame. It is shown in Fig. \ref{BlockDia}(b). Here the relative growth of fire over recent past of the video stream is considered. The average region of the fire flame throughout the video stream and over a few recent frames of video stream are being computed in this part. The fire threat index ($\mathcal{T_F}$) is then quantified with these information. This part is explained in Section \ref{DTF}. 

\section{Formulation of Q-Rough Sets}\label{FQR}
This section discusses the formulation of Q-rough sets. Prior to that formulation, the basics of rough set theory and Q-learning have been briefly described.

\subsection{Rough Sets}
Let there be an information system $S=(\mathcal{U},A)$, where $\mathcal{U}$ is the universe and $A$ is the set of attributes. For any set $B\subseteq A$, there is an equivalance relation $IND(B)$ such that $IND(B)=\{(x,y)\in \mathcal{U}^2|\forall p\in B, p(x)=p(y)\}$, where $p(x)$  function returns the value of the attribute $p$ for data point $x$. The relation $IND(B)$ is called B-indiscernibility relation and any two points $(x,y)\in IND(B)$, i.e., satisfying the B-indiscernibility relation, indicate that $x$ and $y$ can not be distiguishable using the attribute set $B$. Let the equivalence class of B-indiscernibility relation be denoted by $[x]_B$, and $\mathcal{U}|B$ denotes all such equivalence classes. Here $[x]_B$ is called a 'granule' around the data point $x$, created by B-indiscernibility relation. (As stated before, a granule is a clump of object which can not be discriminated with a given attribute set.) Let us denote this granulated information system with $S_B=(\mathcal{U},A,[x]_B)$.

Let $X$ be a set in the universe $\mathcal{U}$ ($X\subseteq \mathcal{U}$) to be approximated based on the equivalence classes $[x]_B$ (i.e., granules) defined over $B$.
Then, $X$ can be
approximated in terms of granules from inner and outer sides as \emph{B-{lower approximation}} $\underline{B}X$ and
\emph{B-{upper approximation}} $\overline{B}X$ respectively. They are
defined as follows:
\begin{equation}\label{UPAPP}
\underline{B}X=\{x\in \mathcal{U} : [x]_{B}\subseteq X\}
\end{equation}
\begin{equation}\label{LOWAPP}
\overline{B}X=\{x\in \mathcal{U} : [x]_{B}\cap X\neq \emptyset\}
\end{equation}

$\underline{B}X$ represents the granules definitely belonging to $X$, while $\overline{B}X$ means granules definitely and possibly belonging to $X$. That means all the elements in $\underline{B}X$ can be certainly classified as member of $X$ on the basis of the knowledge in $B$, while some objects in $\overline{B}X$ can only be classified as possible members of $X$ on the basis of $B$.

\subsection{Q-Learning}

Let the state and action in a given environment be denoted as $s$ and $a$ respectively, and the $Q$-value of doing action $a$ in state $s$ be denoted as $Q(s,a)$. The relation between the direct utility of the state ($U(s)$) and Q-value is as follows.

\begin{equation}\label{DUQ}
    U(s)=\max_aQ(s,a)
\end{equation}
If the Q-values are computed correctly, the following equation (Eqn. \ref{EQI}) will then only reach to an equilibrium. That is, $LHS=RHS$, iff Q-value is correct in Eqn. (\ref{EQI}).
\begin{equation}\label{EQI}
    Q(s,a)=R(s)+\gamma \sum_{s'}P(s'|s,a)\max_{a'}Q(s',a')
\end{equation}
In Eqn. \ref{EQI} $R(s)$ represents the reward value associated with the state $s$; $\gamma$ is the discount factor that determines the importance of future rewards; $P(s'|s,a)$ determines the probability of reaching to the state $s'$ from state $s$ given action $a$, it is determined from the state-transition model; and $Q(s',a')$ is the $Q$-value of future action $a'$ in future state $s'$. As we can see from Eqn. \ref{EQI} an estimated state transition model is required here, therefore, it is not completely model-free. The temporal difference approach or TD Q-learning is the actually model-free one, i.e., state-transition model is not required here. The Q-value updation in TD Q-learning is carried out as follows.
\begin{equation}\label{TDQ}
    Q(s,a)\leftarrow Q(s,a)+ \alpha(R(s)+\gamma (Q(s',a')-Q(s,a))
\end{equation}
$\alpha$ is the learning rate in Eqn. \ref{TDQ}.
\subsection{Q-Rough Set: Definition}
Here in this article we have focused on unsupervised estimation of a set. The method is envisioned to be completely unsupervised and only some basic set of features of the set can be provided apriori. Let us assume the set to be estimated be denoted as $X: X\in \mathcal{U}^2$. The given set of features be $B: B\subseteq A$, where $A$ is the complete set of features. The lower and upper approximations of the set: $\underline{B}X$ and $\overline{B}X$ can now be estimated according to Eqns. \ref{UPAPP} and \ref{LOWAPP} respectively. However, the exact region of the set cannot be estimated through these equations. To extract the exact set and minimize the vagueness in the approximation here we employ a Q-learning agent. Here the set $\underline{B}X$ is used to determine the value function, policy and the state-transition model in Q-learning. The states ($s$) are treated to be the clump of similar data points (granules $[x]_B$) present in the set $\overline{B}X$, the set of actions, is of cardinality equals to 2 and it involves: \textit{$a=\{$updation of $\underline{B}X:\rightarrow \underline{B}X\cup [x]_B$ and moving to the next connected boundary granule to $[x]_B$, keeping $\underline{B}X$ the same and moving to the next boundary granule connected to $\underline{B}_X$$\}$}.

The reward function $R(s)$ is chosen based on the similarities between $[x]_B~s$ and $\underline{B}X$ over $n$-dimensional feature space. The similarities ($D$) between the state ($s\equiv[x]_B$) and model ($M\equiv\underline{B}X$) and reward for that granule ($R([x]_B)$) are now defined as per the following equations.

\begin{eqnarray}\label{qr}
&& D(M,s)=||M-s||_n\\
&& R([x]_B)=1-2 D(M,s).
\end{eqnarray}

In Eqn (\ref{qr}) $R([x]_B)$ is defined in such that the value of the reward should be within the range of $[-1,1]$. Maximum reward (+1) will be generated upon finding total similarity and minimum reward (-1) will be generated upon finding no similarities. 

While determining the Q-value of a boundary granule ($s\equiv[x]_B$), it can be observed that, here the state-action model will lead to two different future states ($s'\equiv[x]_B'$) based on two different actions. But the there is a single possible state for a particular action. Let $[x]_{B1}'$ and $[x]_{B2}'$ be the two possible future states that the agent can reach on taking the actions $a_1$ and $a_2$ respectively. Therefore we are going to get a simple binarized state transition model as follows.
\begin{eqnarray}\label{STM}
P([x]_{B1}'|[x]_B,a1)=1 \nonumber\\
P([x]_{B2}'|[x]_B,a1)=0 \nonumber\\
P([x]_{B1}'|[x]_B,a2)=0 \nonumber\\
P([x]_{B2}'|[x]_B,a2)=1
\end{eqnarray}

Therefore, Eqn. (\ref{EQI}) reduces to Eqn. (\ref{EQIMod}) as follows.

\begin{equation}\label{EQIMod}
    Q([x]_B,a)=R([x]_B)+\gamma\max_{a'}Q([x]_B',a')
\end{equation}

The action, that is maximizing the $Q$-value will be selected. Upon completion of the same process over all the boundary granules, the final set $\underline{B}X$, that we are getting is named as Q-rough approximation of the set $X$. This is how the uncertainty present in the rough approximation of the set $X$ can be minimized, and exact is found out in the incomplete knowledge-base.

\section{Q-Rough Set in Fire Detection}\label{QRSFD}
\begin{figure}
    \centering
    \includegraphics[width=5in]{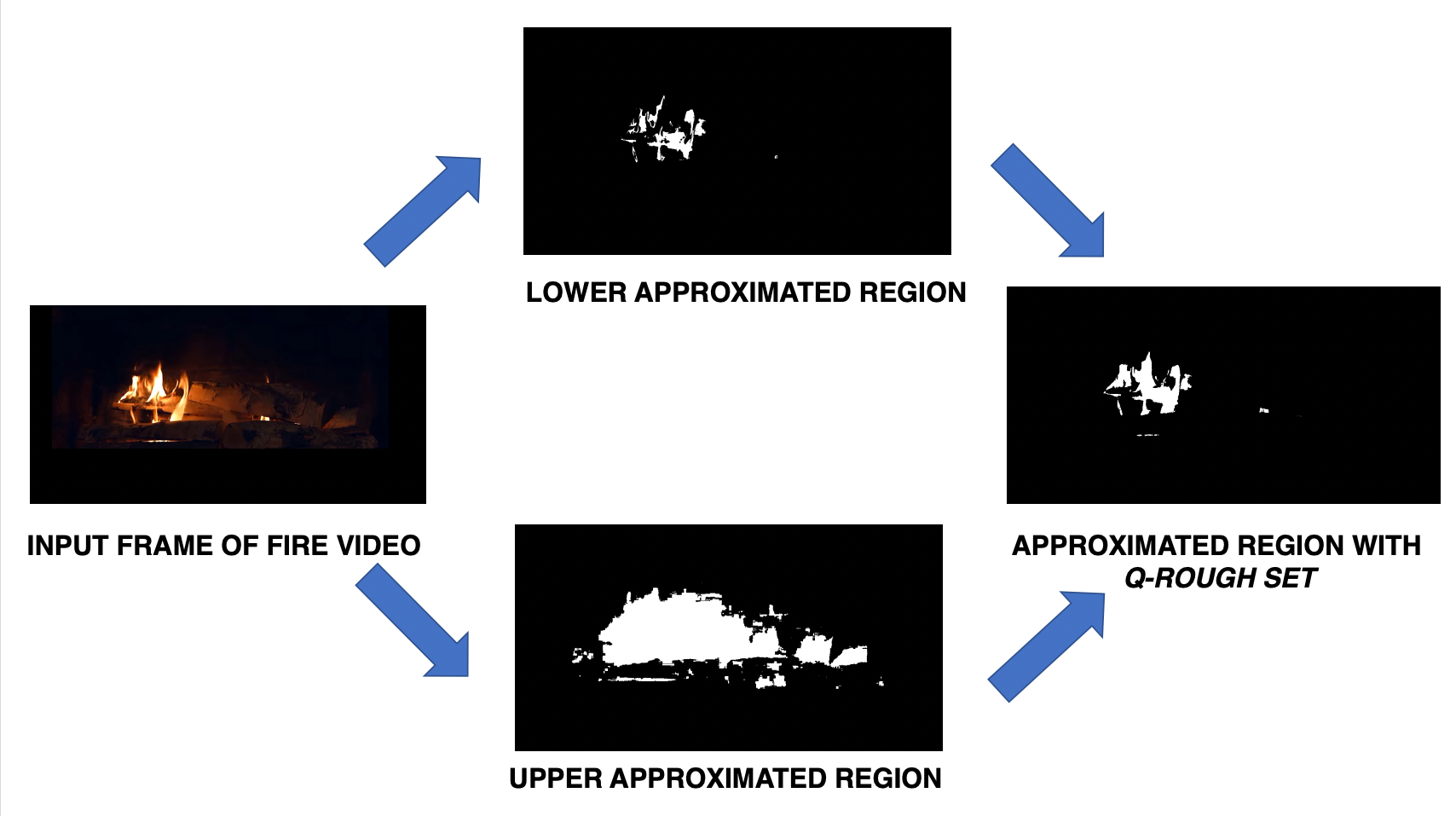}
    \caption{Pictorial Representation of Fire Segmentation with Q-Rough Set}
    \label{fire-seg}
\end{figure}

Let the fire region $F$ be approximated in the input frame $f_t$, given a set of features ($B:B\in A$). We have used the set of rules as defined in \cite{Chen_04}, for R-G-B and Y-Cr-Cb color space to segment out initial fire regions from a video frame. Since our method is unsupervised, we are initially approximating the fire-region in a video frame with these rules. The lower approximation of fire region is done deploying the information of segmented regions in Y-Cr-Cb color space and the upper approximation is done considering the information of segmented region in RGB-color space on the top of the lower approximated region. The final approximation of the fire region in a frame is carried out with Q-rough set. A visual illustration of fire segmentation with Q-rough set is shown in Fig. \ref{fire-seg}. The process of set formation and approximation are shown in the following sections. 

\subsection{Formation of Granules}\label{FGr}
Formation of proper granules plays an important role in approximation in decision making systems. Here while defining Q-rough set for fire segmentation we aim to granulate the image frame based on spatio-color similarities. This is how the separation will be closer to the natural ones. Therefore, we have used spatio-color granulation technique defined in \cite{Pal_17}. That is, a granule around ($\aleph$) an point $p_i$ in the frame $f_t$ is formed according to Eqn. (\ref{clspgr3}).
\begin{align}\label{clspgr3}
\aleph_{sp-clr}(p_i)&=\cup p_j\in U
\end{align}
where $ p_i \quad and \quad p_j$ are binary connected over the
condition,
 $|RGB(p_j)-RGB(p_i)|<Thr$.
 The upper and lower approximations of the fire region ($F$) are carried out over these granules.

\subsection{Lower Approximation of Fire-Region in a Frame}\label{LAF}
 The lower approximated region will consume the segmented region in Y-Cr-Cb color space. The rule-base mentioned in \cite{Zhao_11} for fire segmentation is used here in Y-Cr-Cb. As we know, in Y-Cr-Cb color space \cite{Reinhard_05},  ‘ Y ’ represents the luma component(very similar to the grayscale conversion of the original image ), whereas ‘ Cb ’ and 'Cr' represent chroma components of blue-difference red-difference respectively. The rule-base that is followed for fire segmentation is in Y-Cr-Cb feature space is as follows.

A pixel $p_i$, with Y-Cr-Cb values as $Y(p_i)$, $Cr(p_i)$ and $Cb(p_i)$ will be treated as a fire pixel if it follows any of the following rules.
\begin{itemize}
    \item \textit{Rule 1}: $Y(p_i)\geq Cb(p_i)$ and $Cr(p_i)\geq Cb(p_i)$
    \item \textit{Rule 2}: $Y(x)>Y_{mean}$ and $Cr(p_i)>Cr_{mean}$ and $Cb(p_i)>Cb_{mean}$
    \item \textit{Rule 3}: $Cb(p_i)\leq 120$ and $Cr(p_i)>150$
\end{itemize}
$Y_{mean}$, $Cr_{mean}$, and $Cb_{mean}$ are the mean values of Y, Cr and Cb on that image frame. Let, $F_{YCrCb}$ be the fire-segmented region in the frame $f_t$. Then the lower approximation of the fire region ($\underline{B}F$) is defined according to Eqn. (\ref{FL}). This results in spatio-color granules that are a subset of $F_{YCrCb}$ are considered to be in the lower approximation of the set $F$.
\begin{equation}\label{FL}
    \underline{B}F=\{\aleph(p): \aleph(p)\in F_{YCrCb}\}
\end{equation}

\subsection{Upper Approximation of Fire Region in a Frame}\label{UAF}
The upper approximated region will consume both the segmented outputs of Y-Cr-Cb feature space and RGB feature space. The rules that are defined in \cite{Chen_04,Celik_10,Zhao_11} for fire detection in RGB feature space as as follows. 

A pixel $p_i$ will be detected as fire pixel if it follows the following rules.
\begin{itemize}
    \item \textit{Rule 1}: $R(p_i)\geq R_{mean}$
    \item \textit{Rule 2}: $R(p_i)>G(p_i)>B(p_i)$
\end{itemize}
$R_{mean}$ is the mean value of R over the image frame. Let, $F_{RGB}$ be the segemnted region of fire in RGB feature space. The upper approximated fire region will be derived from $F_{RGB}\cup F_{YCrCb}$ region. The upper approximated fire region ($\overline{B}F$) is defined as according to Eqn. (\ref{FU}).
\begin{equation}\label{FU}
    \overline{B}F=\{\aleph(p): \aleph(p)\cap (F_{RGB}\cup F_{YCrCb})\neq \emptyset\}
\end{equation}
\subsection{Computation of Reward Function}
Proper computation of reward function is another major concern while defining Q-rough sets. As defined in Eqn. (\ref{qr}), the distance between the lower approximated region and a boundary granule is to be computed to determine the reward function. We know that the no. of data points in a $\underline{B}F$ and a boundary granule $\aleph(p)$ can never be the same. Therefore, instead of computing point-to-point distance between these two sets, we are computing the distance between their mean values in different feature space. Therefore, Eqn. (\ref{qr}) can be re-written here as follows. \begin{eqnarray}\label{qrF}
&& D(\underline{B}F,\aleph(p))=||mean(\underline{B}F)-mean(\aleph(p))||_{RG BYCrCb}\\
&& R(\aleph(p))=1-2 D(\underline{B}F,\aleph(p)).
\end{eqnarray}
In Eqn. (\ref{qrF}) normalized Euclidean Distance is considered as a distance metric. The Q-values will now be updated according to Eqn. (\ref{EQIF}).

\begin{equation}\label{EQIF}
    Q(\aleph(p),a)=R(\aleph(p))+\gamma\max_{a'}Q(\aleph(p)',a')
\end{equation}

\subsection{Algorithm for Fire Detection with Q-Rough Set}
We are going to describe the proposed method step-wise for fire detection from a video frame $f_t$ with Q rough set. The theoretical details are presented in the preceding sections. The detailed methodology is summarized in the form of an algorithm in Algorithm \ref{algSPO}. 
\begin{algorithm}
 \caption{Fire Detection from A Frame with Q-Rough Set}          
\label{algSPO}                           
\begin{algorithmic}                    
    \STATE INPUT: $f_t$
    \STATE OUTPUT: $f_t$ with segmented fire region $F$
    \STATE INITIALIZE: $\underline{B}F=\overline{B}F$ $\Leftarrow\emptyset$
    \item 1: Granulate the frame $f_t$ as described in Section \ref{FGr}
    \item 2: Segment out $F_{YCrCb}$ and $F_{RGB}$ as described in Sections \ref{LAF} and \ref{UAF} respectively.
    \item 3: Define $\underline{B}F$ and $\overline{B}F$ following Eqns. (\ref{FL}) and (\ref{FU}).
    \item 4: For a boundary granule $\aleph{p}\in\{\overline{B}F-\underline{B}F\}$ do the following.\\
    ~i) Compute $R(\aleph(p))$ with Eqn. (\ref{qrF})\\
    ~ii) Compute $Q(\aleph(p),a_1)$ and $Q(\aleph(p),a_2)$, with Eqn. (\ref{EQIF})
    \IF{$Q(\aleph(p),a_1)>Q(\aleph(p),a_2)$}
    \STATE Set $\underline{B}F=\underline{B}F\cup \aleph(p)$ and move to the next connected granule of $\aleph(p)$
    \ELSE
    \STATE Remove $\aleph(p)$ from from $\overline{B}X$ and moving to the granule connected to $\underline{B}X$. 
    \ENDIF
    \item 5: Repeat Step 4 for the next boundary granule.
    \item 6: Repeat Step 4 and 5 till $\{\overline{B}F-\underline{B}F\}=\emptyset$
    \item 7: Set $F=\underline{B}F$
    \end{algorithmic}
\end{algorithm}

\subsection{Fire Segmentation Over The Video Sequence}
So far we have discussed the method of fire segmentation with Q-rough sets over a video frame. Since we are dealing with video sequence, we do not need to repeat the entire process for each frame. Rather, we will employ our trained Q-agents to explore the fire region in the upcoming frames. That is, the Q-agents are going to check the spatio-color granules and in rest of the sequence, and similar granules will automatically be included in the lower approximated fire region.

Let fire region in the $t^{th}$ frame ($f_t$) be approximated as $F_t$ with Algorithm 1. Let $\aleph(p_{t+1})$ be a granule under consideration in the frame $f_{t+1}$. The action over $\aleph(p_{t+1})$ will get automatically decided by the Q-agent in $f_{t+1}$, since it already knows which action maximizes the utility from $f_t$. The computation of Q-value will only be required in $f_{t+1}$ if $\aleph(p_{t+1})$ is an unknown state to the Q agent.

\section{Determination of Threat of Fire}\label{DTF}
After determination of fire regions in video sequences, we focus to determine the the threat associated with the fire. Fire is a necessity as well as a threat in human civilization. Fire is used for light, for heat, to cook etc. But fire is desired to be used keeping it in control, once it becomes uncontrolled, it creates threat and damages. Here we plan to develop a methodology to determine whether the fire is in use or it is becoming a threat. We have observed that, fire is generally in a controlled position while the flames are within a certain space, but it becomes a threat when the flame starts to spread. The faster the flames get spread, greater is the threat. Here we aim to quantify this phenomenon and accordingly generate an alarm with the possible threat. 

Here we have estimated relative spread in fire regions in a video sequence to quantify the threat. If the flame is in control, the fire regions may change from frame to frame, but there will be a flicker effect. That is, the fire regions may remain within some limits throughout the sequence. But if the fire becomes uncontrolled, the fire region will spread rapidly compared to its previous frames. This is why relative spread is considered for the threat computation. Let $F_t$ be the segmented fire region in frame $f_t$. Let the information from $P$ number of previous frames be used to determine the threat. The average fire segments ($F_{\mu}$) throughout the sequence and the and recent average fire segments ($F_{\mu P}$) are computed based on the following equations. 
\begin{eqnarray}\label{thseg}
&& F_\mu=\frac{\sum_{i=1}^N F_i}{N}\\
&& F_{\mu P}=\frac{\sum_{i=N-P}^N F_i}{P}: P<N
\end{eqnarray}
The threat index of fire ($\mathcal{T_F}$) is now defined with relative increment of spread in recent frames. It is computed with Eqn. (\ref{thf}).

\begin{equation}\label{thf}
   \mathcal{T_F}=\frac{F_{\mu P}-F_\mu}{F_\mu} 
\end{equation}

Please note that a signed difference is considered in the numerator of right hand side of in Eqn. (\ref{thf}). This is because, the threat will only be positive if there if a relative spread in fire region, if the region is concentrated or the fire is extinguishing, the threat will become negative. Thereby eliminating occurrences of false alarms. 

Selection of the value of $P$ plays a major role in this index. Here we have decided to consider the relative growth of fire over the last one second. Therefore, $P$ is chosen based on the frame rate of video acquisition per second (fps). That is, if the video is acquired with a frame rate of 30 fps, $P$ is set to be $30$.

\section{Experimental Results}\label{res}
Here in this section we have experimentally demonstrated the effectiveness of the proposed methods in identifying the followings. The unsupervised fire segmentation method, developed here is proven to be superior in terms of qualitative and quantitative results over different types of input videos, from, a few state of the art methods. The threat detection method is also found to be effective in quantifying the threat associated with the fire. 

We have conducted our experiment with 30 different types of videos including more than 25000 video frames, with fire flame. Ten video sequences out of the thirty considered videos contain spreading fire or uncontrolled fire, and the rest contain controlled fire. Ten video sequence sets were acquired from the sets from 'Firesence' database \cite{Grammalidis_17}, and twenty video sequences from different links freely available on 'YouTube'. To limit the siz of the article we have shown the qualitative and quantative results obtained over seven different video sequences. Out of which we have selected only two videos where the fire region remained to be within a range, fire is either rapidly growing, or slowly spreading in the rest of the five videos shown here. 

The video sequences, over which we have shown the results are described as follows. 'PosVideo1' \cite{Grammalidis_17} shows a part of a bus that is set on fire, the fire is gradually growing here. 'PosVideo2' \cite{Grammalidis_17} shows fire is set into a kitchen and it is growing gradually. In 'PosVideo4' and 'PosVideo5' \cite{Grammalidis_17} fire is under control, two different burning fire-places are captured in these videos. Two men are setting fire in a forest region and fire start to grow is the content of 'PosVideo6' \cite{Grammalidis_17}. In 'WeddingDress' \cite{IE_18} video, fire is initially spreading over the wedding dresses and then becomes constant, whereas in 'ChristmasTree' \cite{NFA_15} video a Christmas tree catches fire that spreads rapidly. The effectiveness of Q-rough sets in segmenting out the fire regions is shown in the next section.

\subsection{Fire Region Segmentation with Q-rough Set}
Here in this section we are going to demonstrate the effectiveness of Q-Rough sets in identifying the fire segments throughout the video sequences. The visual segmented outputs for four different frames over seven different sequences, described above, are shown in Figs. \ref{VisCompM}(1) to \ref{VisCompMcont}(7). It can be concluded from the visual results that the fire regions are well segmented out with Q-rough sets. 

The segmentation accuracy is quantitatively demonstrated in Table \ref{FPFN}. Please note that, there was no ground truth available for the video sequences used here for the experimentations. Therefore, we have manually annotated ten frames, selected randomly from each video sequence, and performed the quantitative analysis. The average $False positive ~(FP)$, $False Negative~ (FN)$, $Precision$, and $Recall$ values are given here in Table \ref{FPFN}.
\begin{table}[ht]
    \centering
    \caption{Fire Segmentation Accuracy of Q-Rough Sets}
    \begin{tabular}{|c|c|c|c|c|}
    \hline
        Sequence & $FP(\%)$ & $FN(\%)$ & $Precision$ & $Recall$\\
         \hline
        PosVideo1 & 0.5 & 15 & 0.99 & 0.87\\
        \hline
PosVideo2 & 3 & 5 & 0.97 & 0.95\\
\hline
PosVideo4 & 4 & 1 & 0.96 & 0.99\\
\hline
PosVideo5 & 3 & 2 & 0.97 & 0.98\\
\hline
PosVideo11 & 2 & 10 & 0.98 & 0.9\\
\hline
WeddingDress & 2 & 8 & 0.98 & 0.91\\
\hline
ChristmasTree & 23 & 5 & 0.77 & 0.93\\
\hline
    \end{tabular}
    \label{FPFN}
\end{table}
From Table \ref{FPFN} it can be observed that, the Q-rough set method is found to be quite efficient in segmenting out the fire regions almost in every video except ChistmasTree. The fire regions are well segmented even the ones that are small in size (PosVideo11 and PosVideo1), continuously burning (PosVideo4 and PosVideo5) or spreading rapidly (PosVideo2 and WeddingDress). In the ChristmassTree video the fire gets reflected in the wall and the wall reflects almost of similar color as that of the fire. Therefore, some part of the wall also gets segmented out as fire. 

\begin{figure}[ht]
\includegraphics[width=5in]{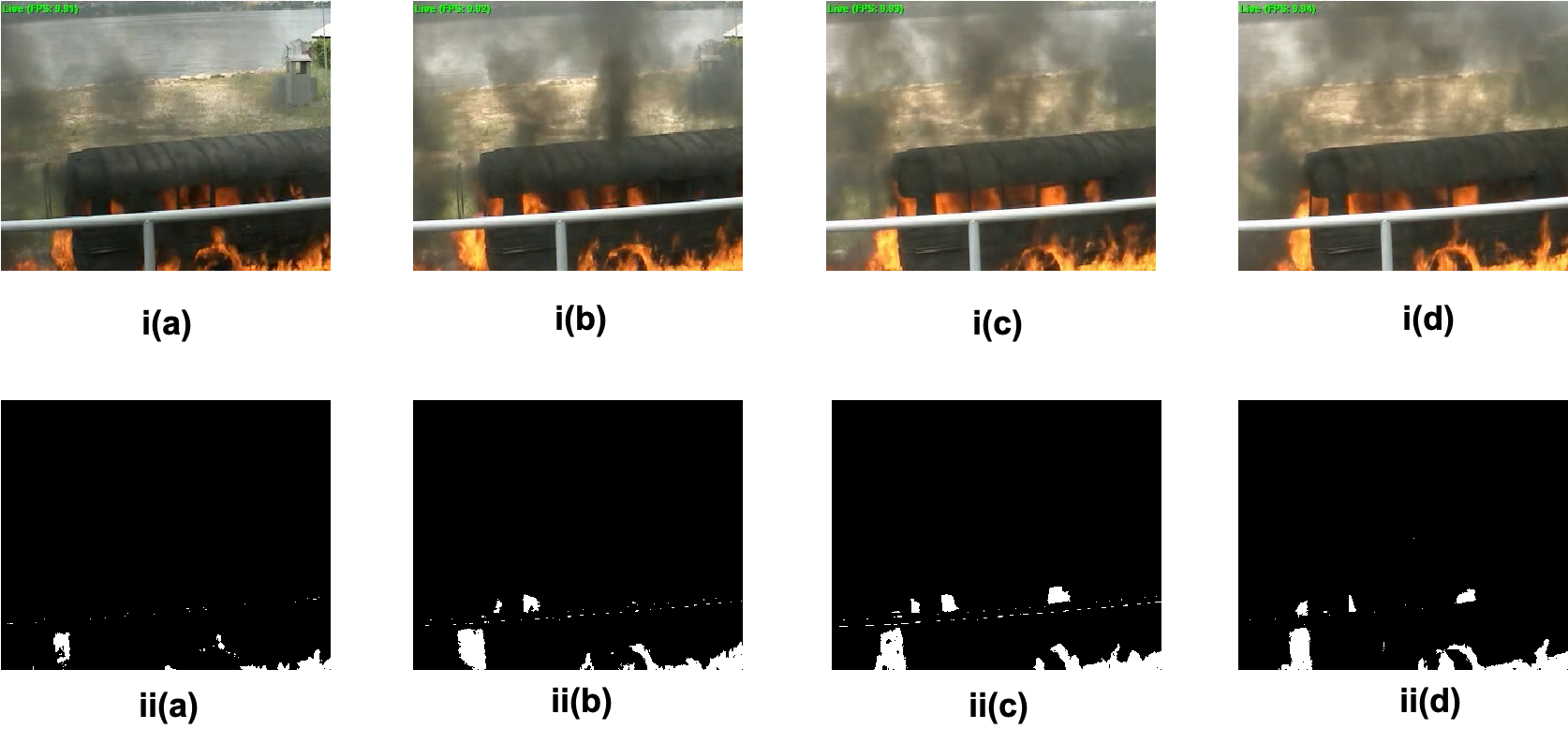}~\textbf{(1)}\\
\includegraphics[width=5in]{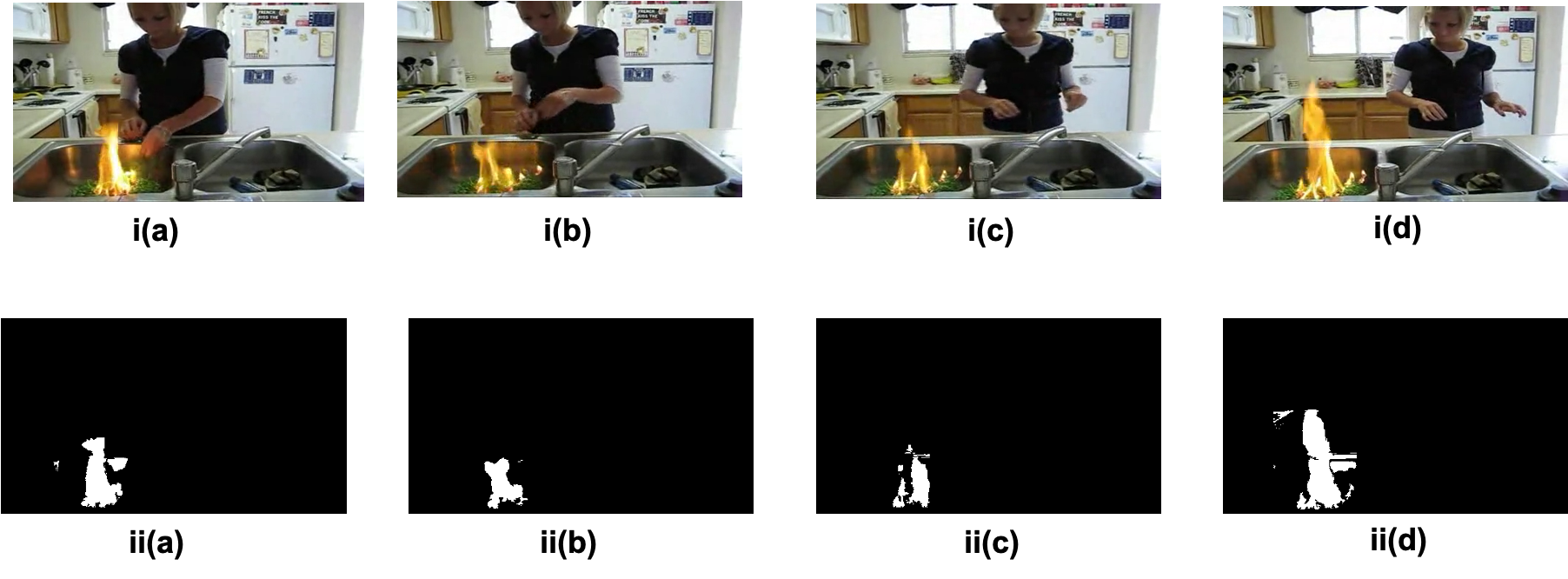}~\textbf{(2)}\\
\includegraphics[width=5in]{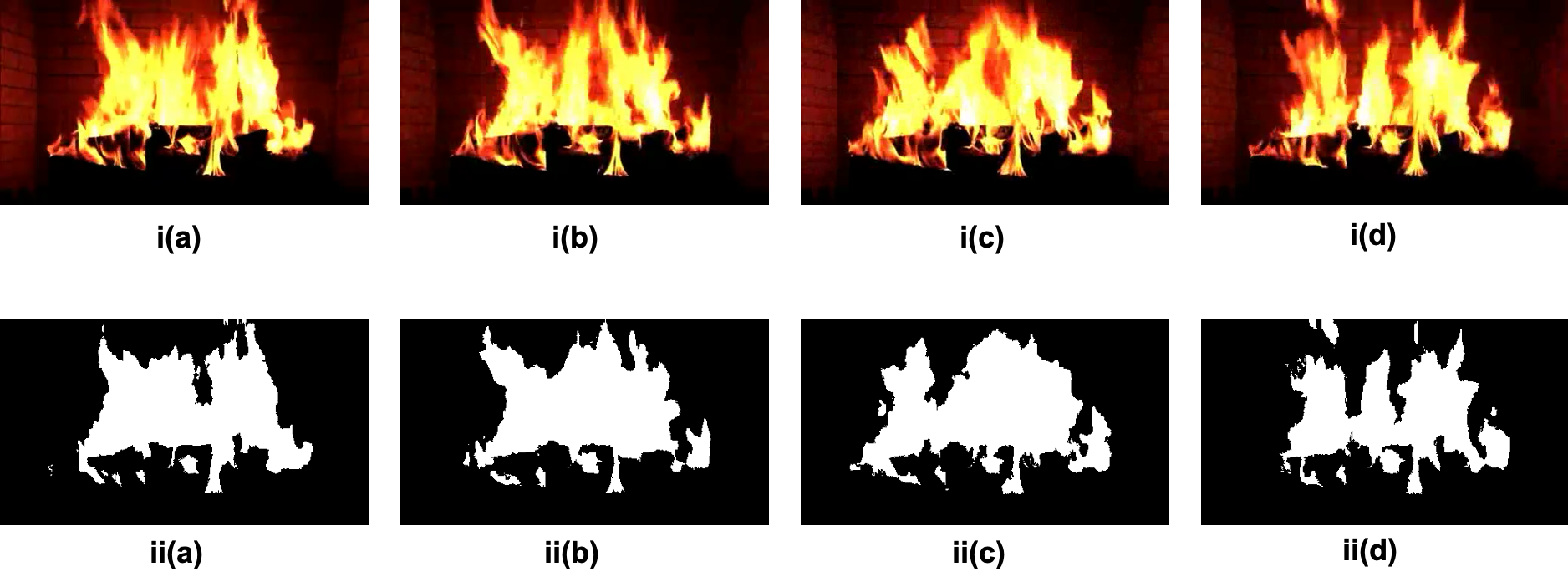}~\textbf{(3)}\\
\includegraphics[width=5in]{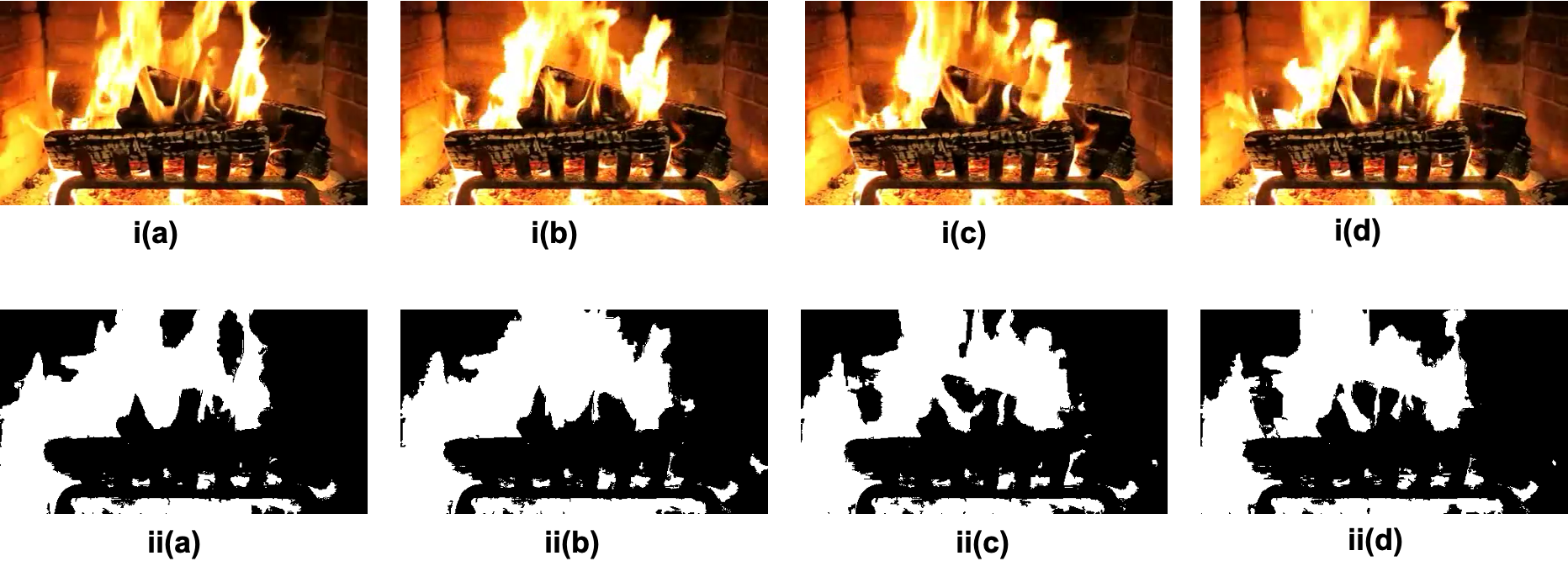}~\textbf{(4)}\\
\caption{{Fire segmentation results with Q-rough sets for frame nos. (1) 15, 45, 75, 105 of 'PosVideo1' sequence (2) 25, 45, 85, 105 of 'PosVideo2' sequence (3) 25, 45, 85, 105 of 'PosVideo4'  sequence,
   (4) 25, 45, 85, 105 of 'PosVideo5'  sequence}}
   \label{VisCompM}
\end{figure}
    \begin{figure}[ht]\ContinuedFloat
\includegraphics[width=5in]{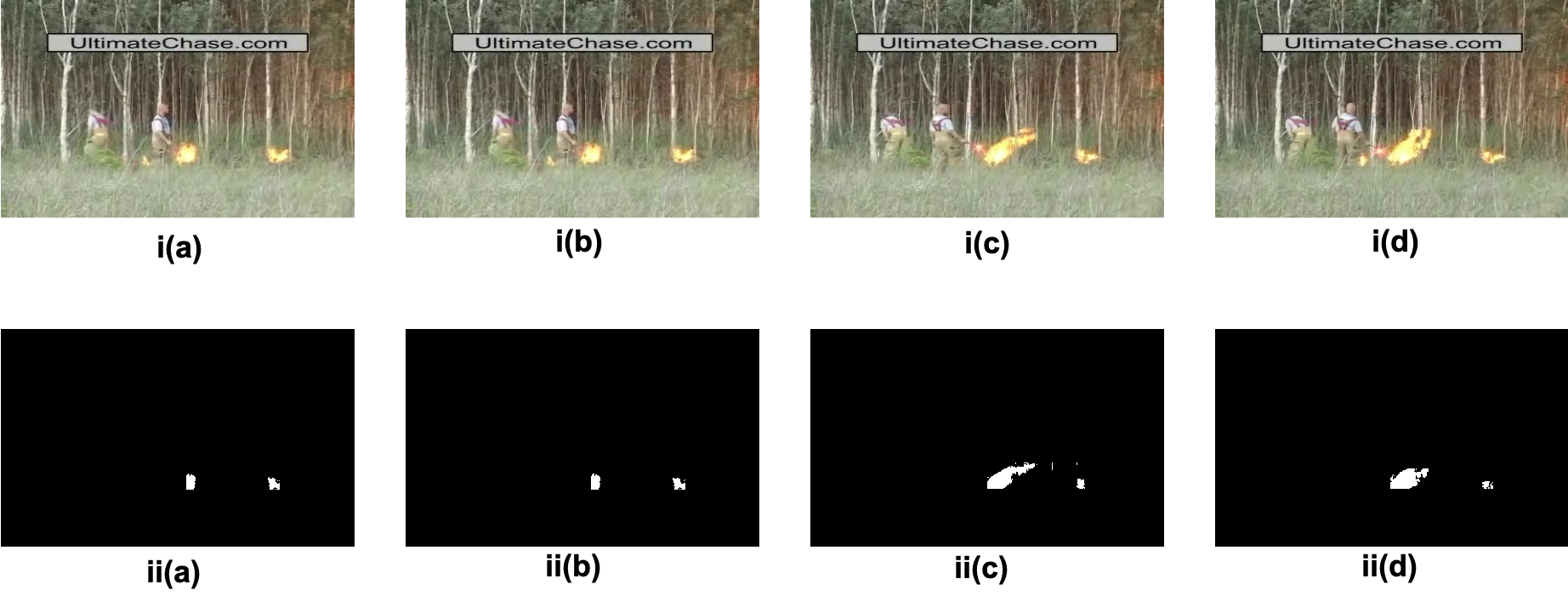}~\textbf{(5)}\\
\includegraphics[width=5in]{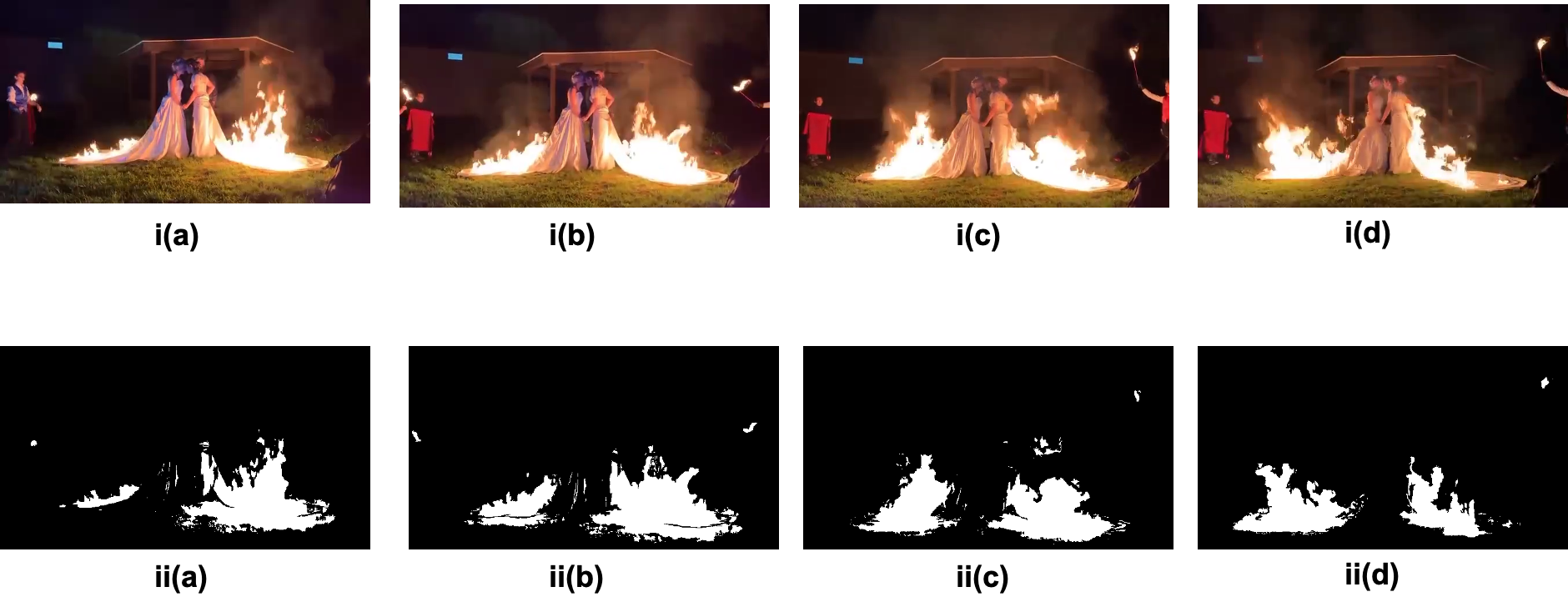}~\textbf{(6)}\\
\includegraphics[width=5in]{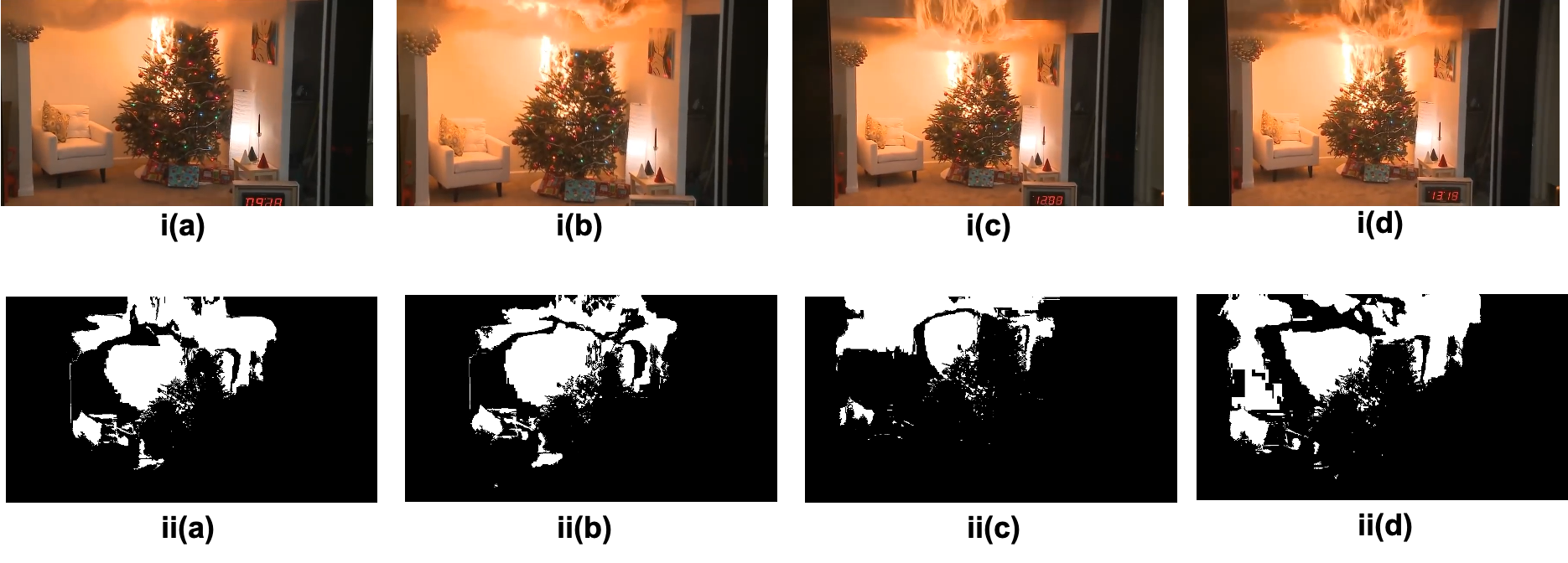}~\textbf{(7)}\\
 \caption{Continuation of Fig. \ref{VisCompM} (5) 25, 45, 85, 105 of 'PosVideo11'  sequence, (6) 25, 45, 85, 105 of 'WeddingDress'  sequence, and (7) 25, 45, 85, 105 of 'CristmassTree'  sequence; (i) input, (ii) segmented}
  \label{VisCompMcont}
\end{figure}

\subsection{Comparative Study}
Here in this section we have demonstrated qualitative and quantitative comparison of our fire segmentation method with four state of the art methods. The methods, with which the comparative studies are performed include: i) RGB and motion feature based method with HMM (RGB-M) \cite{Toreyin_06}, ii) HSV model based method developed for forest fire detection with static and dynamic features (HSV-SD) \cite{Zhao_11}, iii) spatio-temporal flame modelling method (ST-F) \cite{Chino-15}, and iv) fire detection with improved deep learning (F-IDL) \cite{Cai_19} method. 

\begin{figure}[ht]
\includegraphics[width=5in]{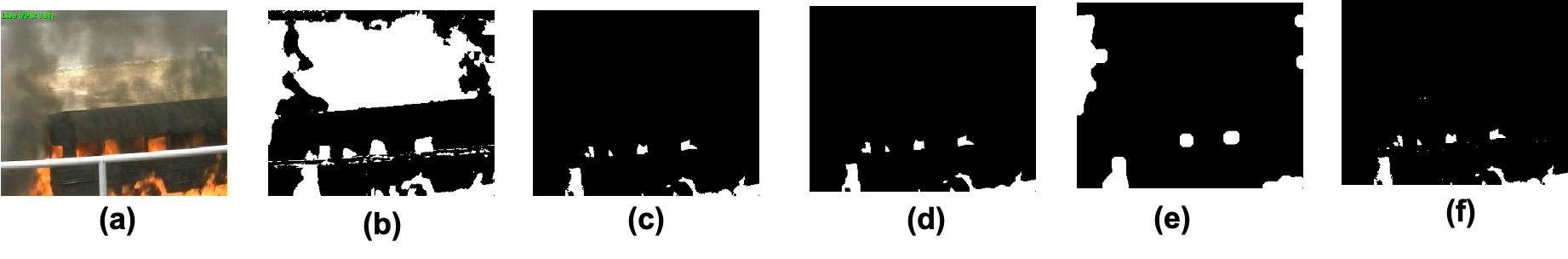}~\textbf{(1)}\\
\includegraphics[width=5in]{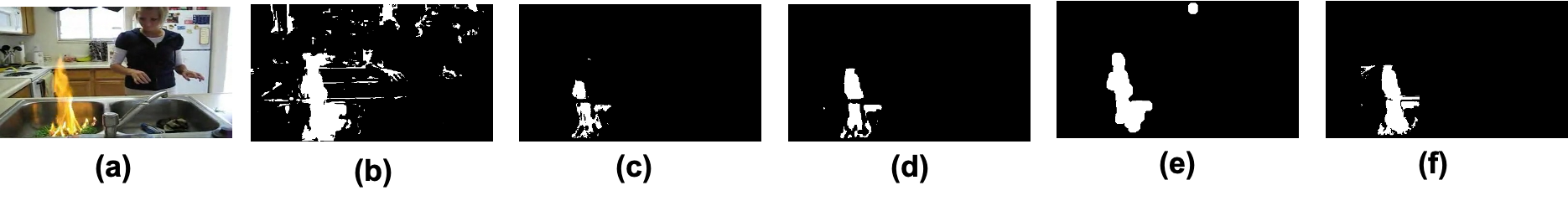}~\textbf{(2)}\\
\includegraphics[width=5in]{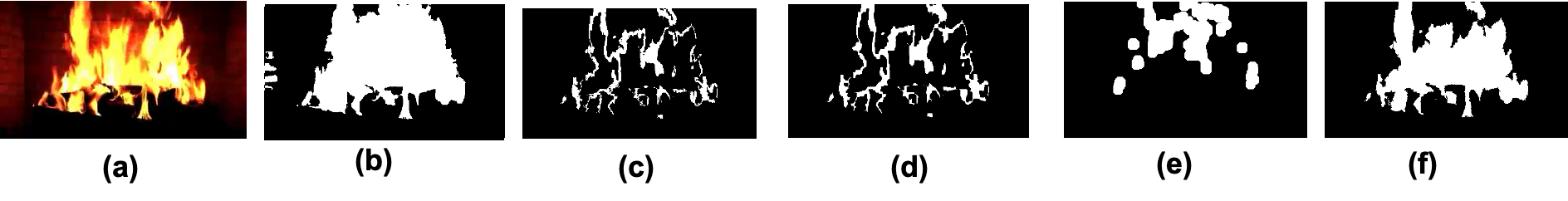}~\textbf{(3)}\\
\includegraphics[width=5in]{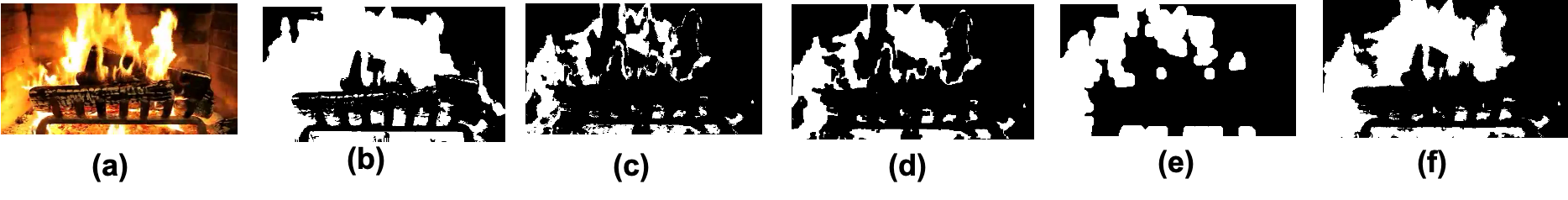}~\textbf{(4)}\\
\includegraphics[width=5in]{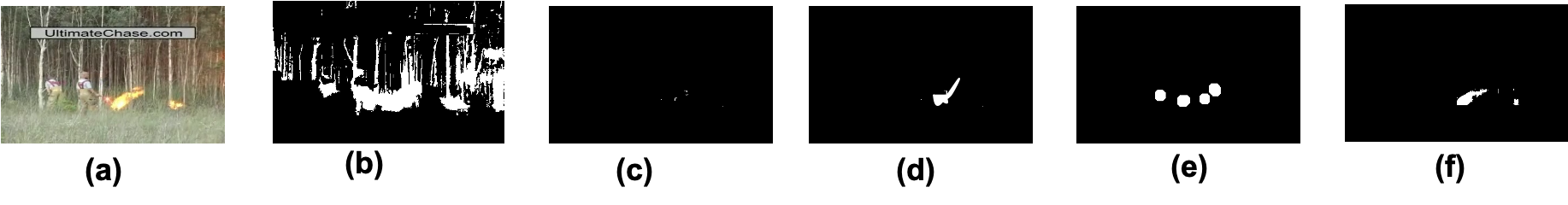}~\textbf{(5)}\\
\includegraphics[width=5in]{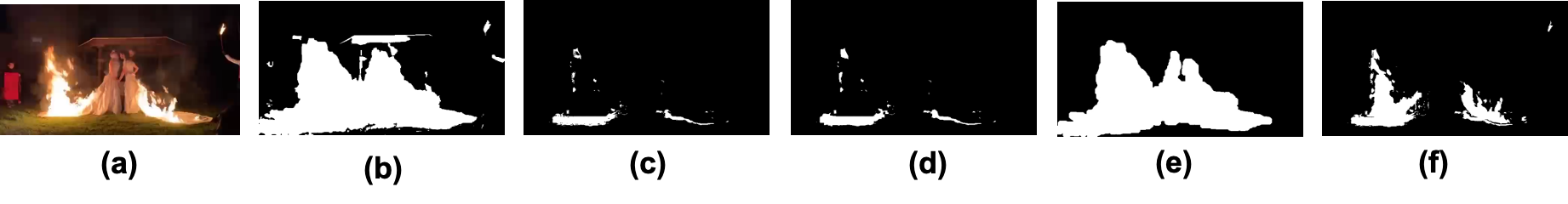}~\textbf{(6)}\\
\includegraphics[width=5in]{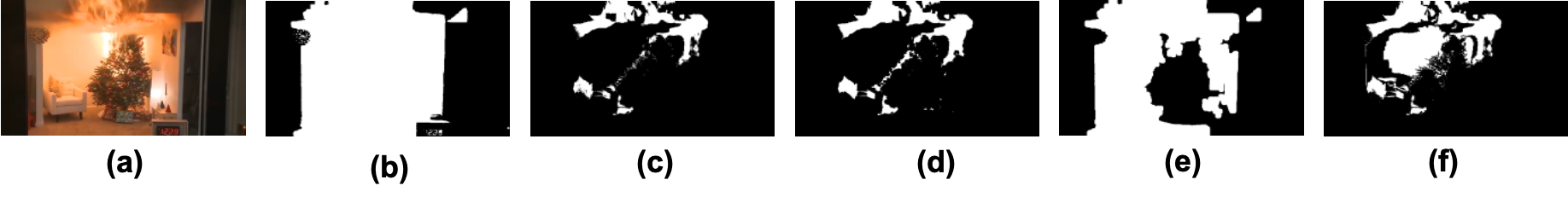}~\textbf{(7)}\\
\caption{Visual comparison on fire segmentation on (1) PosVideo1 sequence, (2) PosVideo2 sequence, (3) PosVideo4 sequence, (4) PosVideo5 Sequence, (5) PosVideo11 sequence, (6) WeddingDress sequence, and (7) ChristmassTree Sequence; (a) input, (b) RGB-M method, (c) HSV-SD method, (d) ST-F method, (e) F-IDL method, and (f) Q-rough set (proposed) method}
   \label{SegComp}
\end{figure}

The visual comparison over a frame of the seven different video sequences on with the five different methods are shown in Fig. \ref{SegComp}. From the visual results it can be seen that the Q-rough set method is segmenting out the fire region most efficiently. 

Please note that, none of the methods used in the comparative study focus on classifying each fire pixel to the non-fire ones, therefore, the measures of classification accuracy, like, true positive, false negative is not going to be a fair parameter here for the comparative study. Therefore, a measure, namely, RMSE (root mean square error) is taken under consideration here for quantitative comparison. The root mean square error between the four corner pixels of the bounding boxes, covering the ground truth regions of fire, and the segmented regions of fire are considered here for the measurement. The average RMSE values for the seven videos sequences with five methods are given in Table \ref{RMSE}

\begin{table}
    \centering
    \caption{Fire Region Segmentation: Comparison with Average RMSE}
    \begin{tabular}{|c|c|c|c|c|c|}
    \hline
        Sequence & $RGB-M$ & $HSV-SD$ & $ST-F$ & $F-IDL$ & $Q-Rough Set$\\
         \hline
        PosVideo1 & 52.3 & 8.2 & 6.6 & 13.4 & 5.8\\
        \hline
PosVideo2 & 46.8 & 11.1 & 7.2 & 5.8 & 6.9\\
\hline
PosVideo4 & 9.2 & 6.3 & 5.9 & 6.1 & 5.5\\
\hline
PosVideo5 & 8.8 & 5.2 & 4.7 & 5 & 4.9\\
\hline
PosVideo11 & 48.6 & 11.2 & 9.5 & 7.3 & 4.9 \\
\hline
WeddingDress & 19.4 & 17.8 & 8.3 & 12.2 & 5.1\\
\hline
ChristmasTree & 34.6 & 13.5 & 12.6 & 21.2 & 16.3\\
\hline
    \end{tabular}
    \label{RMSE}
\end{table}

It can be observed from Fig. \ref{SegComp} and Table \ref{RMSE} that the proposed unsupervised method performs superior or equally well to those of the state of the art methods. Besides, it can also be seen from the visual results (Fig. \ref{SegComp}), that Q-rough set method can classify the fire pixels from the non-fire ones better than the other methods in comparison. 

\subsection{Effectiveness of Fire Threat Index $\mathcal{T_F}$}
\begin{figure}[ht]
\includegraphics[width=2.5in]{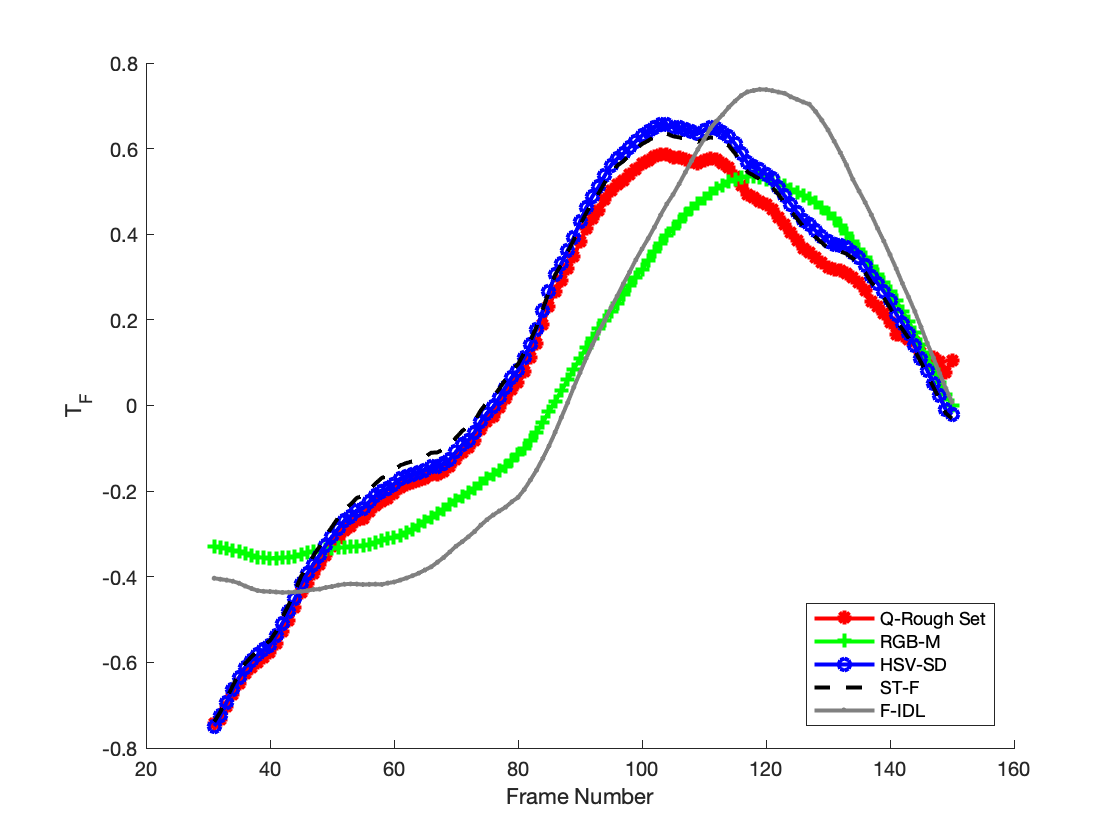}~\textbf{(1)}
\includegraphics[width=2.5in]{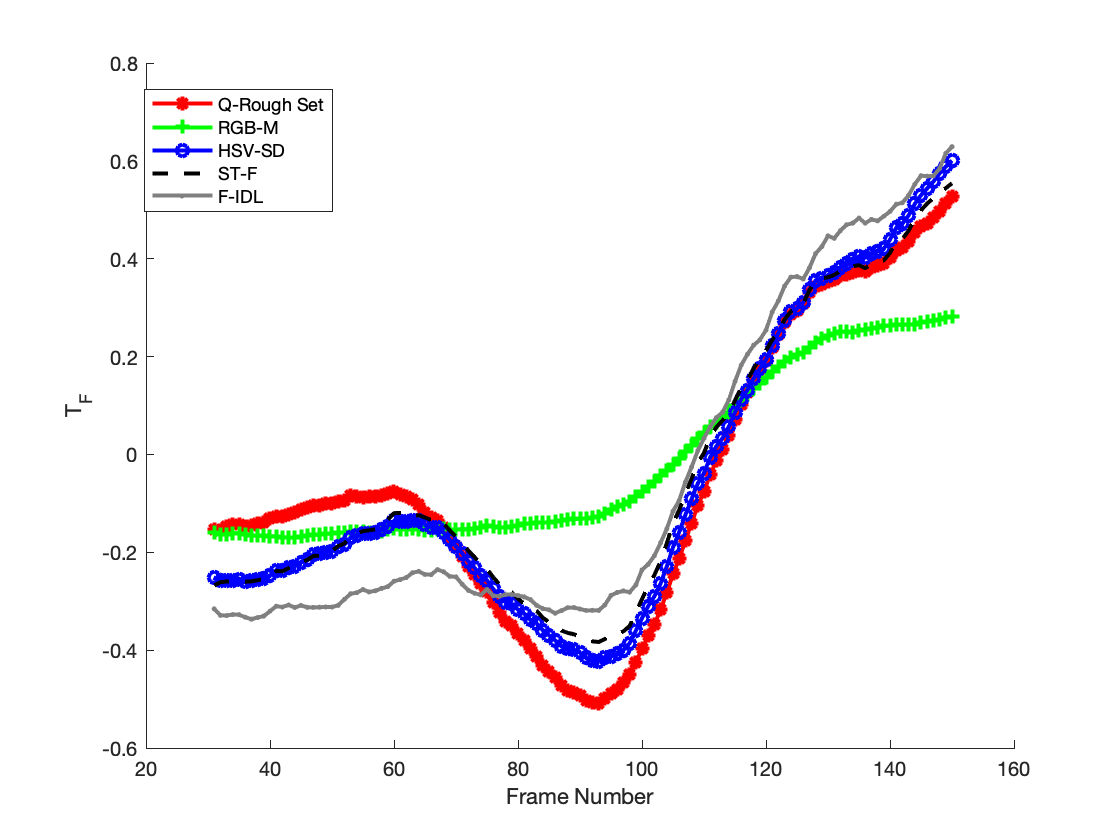}~\textbf{(2)}\\
\includegraphics[width=2.5in]{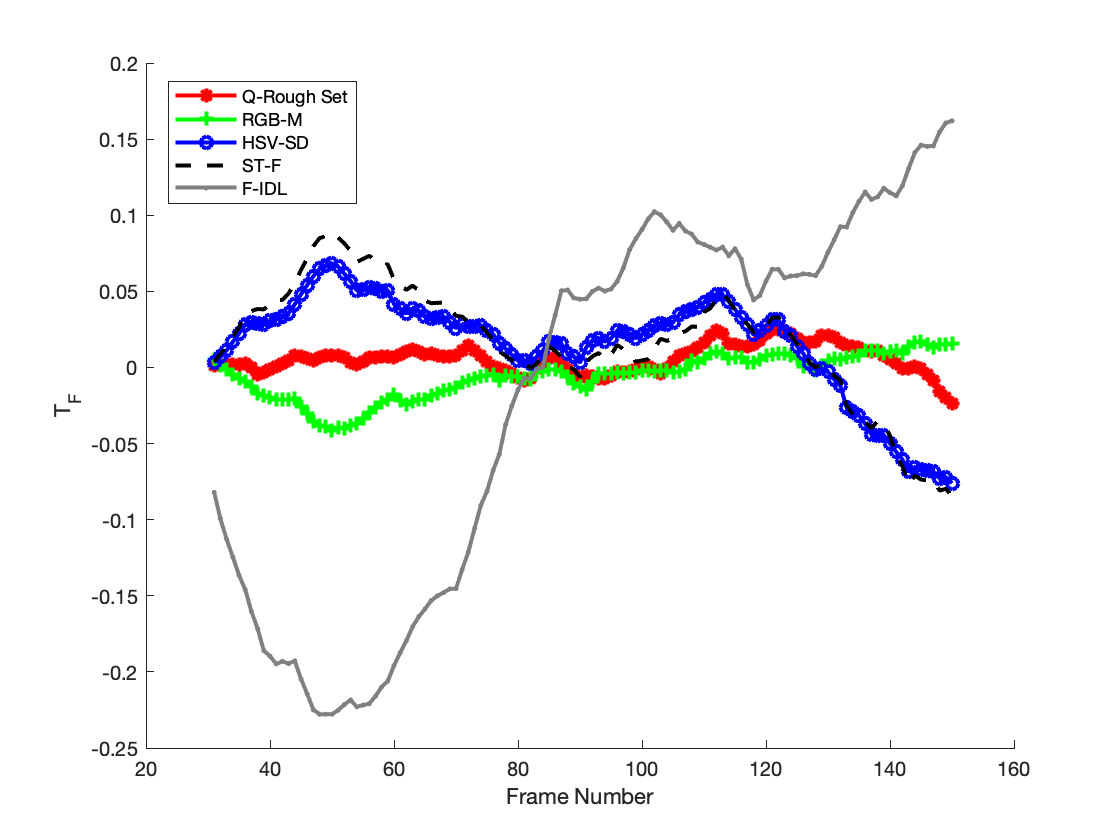}~\textbf{(3)}
\includegraphics[width=2.5in]{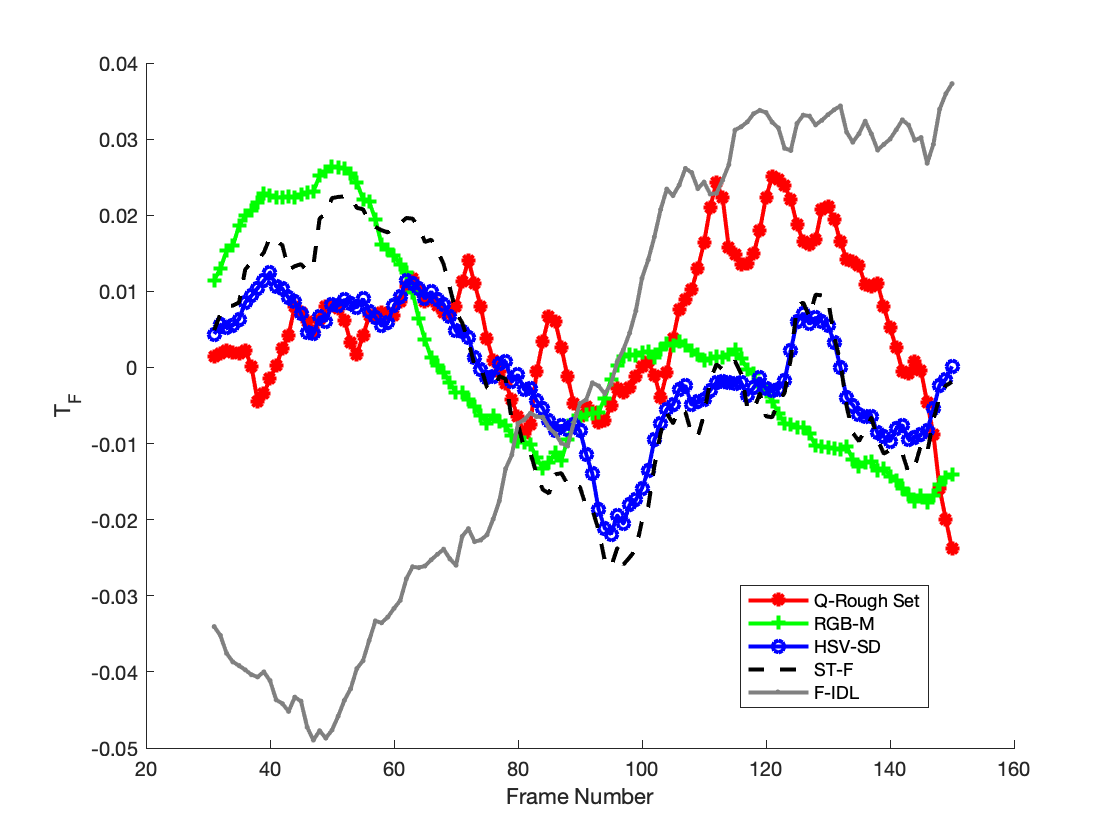}~\textbf{(4)}\\
\includegraphics[width=2.5in]{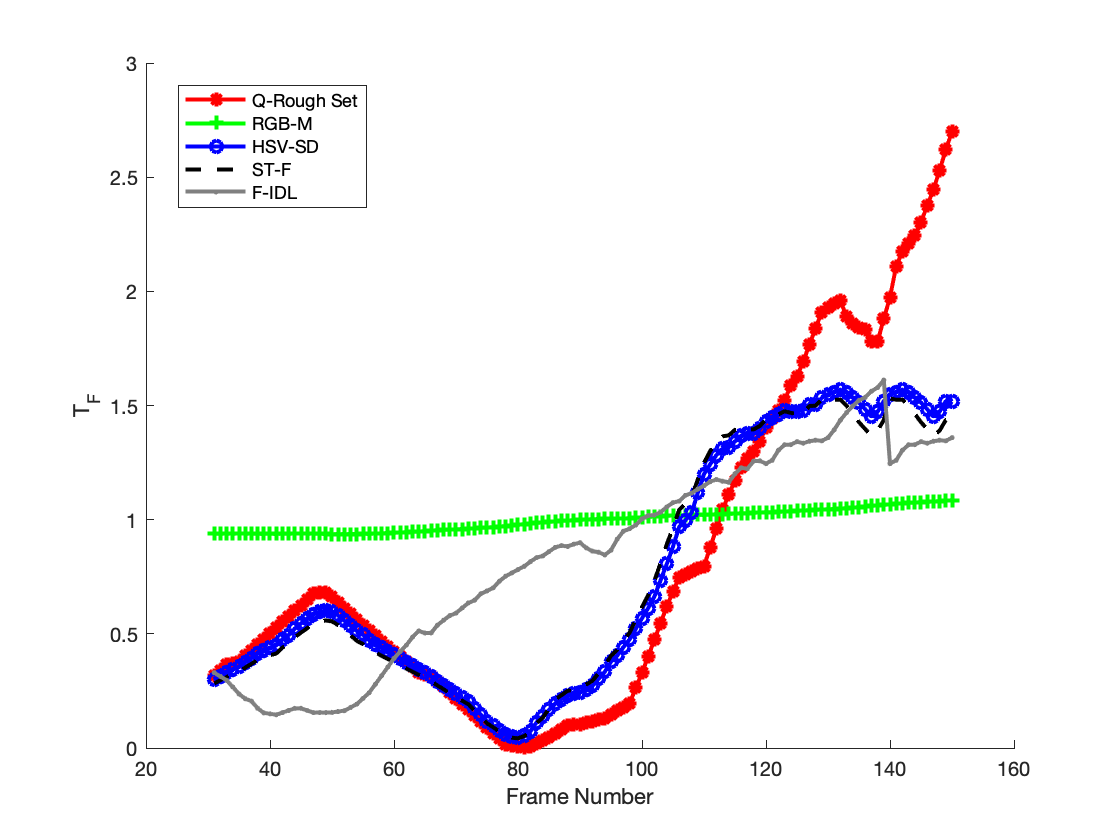}~\textbf{(5)}
\includegraphics[width=2.5in]{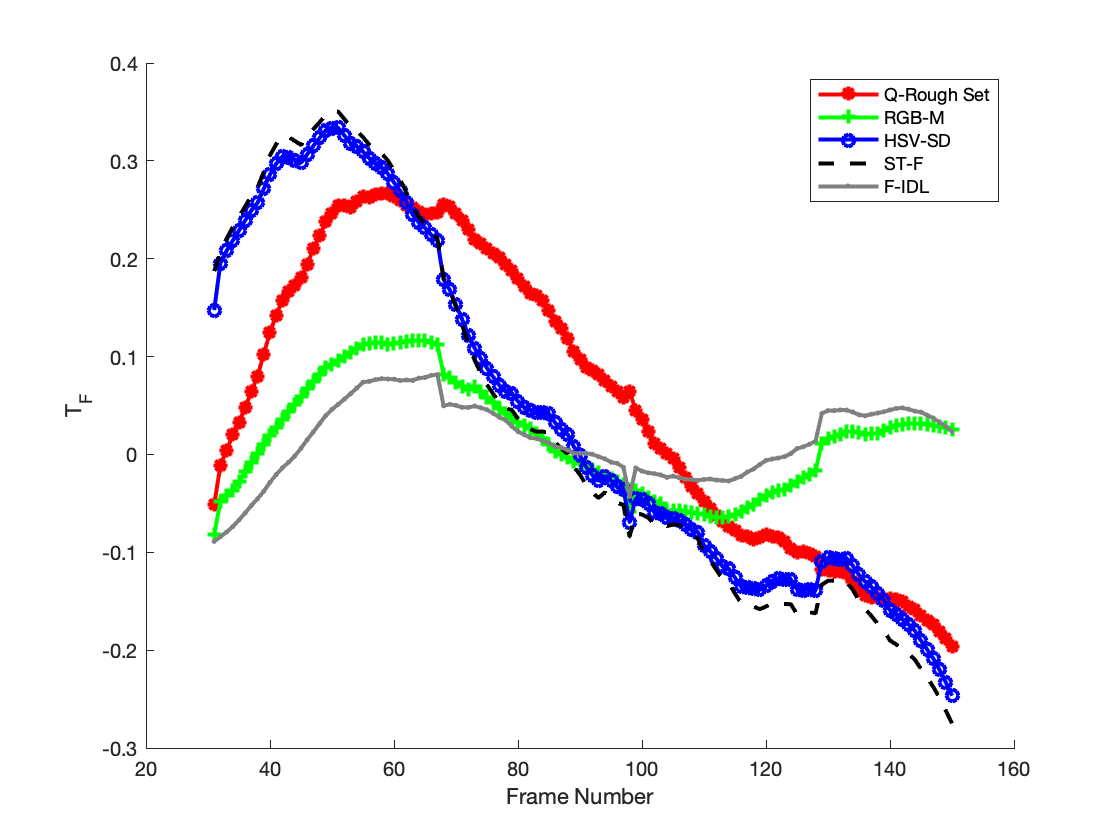}~\textbf{(6)}\\
{\includegraphics[width=2.5in]{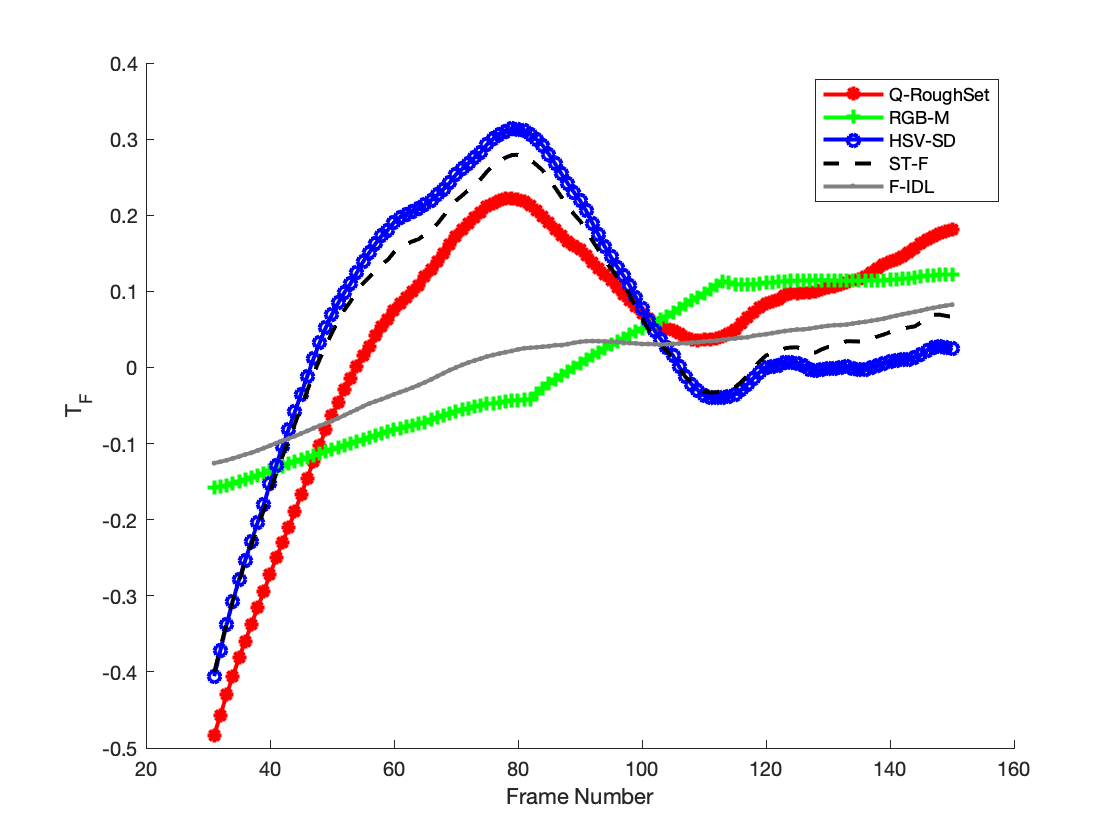}}~\textbf{(7)}\\
\caption{$\mathcal{T_F}$ index values over 1) PosVideo1 sequence, (2) PosVideo2 sequence, (3) PosVideo4 sequence, (4) PosVideo5 Sequence, (5) PosVideo11 sequence, (6) WeddingDress sequence, and (7) ChristmassTree Sequence}
   \label{TFI}
\end{figure}

The effectiveness of $\mathcal{T_F}$ index in identifying the threat of fire is validated here with extensive experiments. The values of $\mathcal{T_F}$ for each frame throughout the seven video sequences are plotted here in Fig. \ref{TFI}. The $\mathcal{T_F}$ values for all the five methods, over which the comparative studies are performed, are computed here and plotted accordingly in Fig. \ref{TFI}. The red line represents the values of $\mathcal{T_F}$ indices obtained by Q-rough set method, the green line indicates $RGB-M$ method, blue line indicated $HSV-SD$ method, gray '- -' line indicates $ST-F$ method, and the solid black line indicates $F-IDL$ method in Fig. \ref{TFI}. The increament or decrement of fire flame is well reflected by the proposed $\mathcal{T_F}$ index. For example, fire is initially increasing and then decreasing in $PosVideo1$-sequence which is reflected in the curve in Fig. \ref{TFI}(1) almost by all the methods. The increasing threat of fire of $PosVideo2$-sequence can be inferred from Fig. \ref{TFI}(2) with the increasing values of $\mathcal{T_F}$. Fire flames have a flicker effect, and no threat in $PosVideo4$ and $PosVideo5$ sequences. It is best reflected by Q rough set method in Figs. \ref{TFI}(3) and \ref{TFI}(4), where the $\mathcal{T_F}$ values are withing a fixed range closed to the value zero. The gradually spreading fire in $PosVideo11$-sequence can be inferred from \ref{TFI}(5) with Q-rough set method. Initial spread and then concentration of fire in $WeddingDress$-sequence is reflected well in Fig. \ref{TFI}(6). The sudden spread of fire in $ChristmasTree$-sequence can be well detected in Fig. \ref{TFI}(7). Therefore, the proposed $\mathcal{T_F}$ index is successful in identifying the threat of fire correctly. 

\section{Conclusions and Future Work}\label{conF}
In this article we have primarily addressed two tasks related to fire threat detection from videos. The first one is unsupervised fire region segmentation with Q-rough sets and the second one is definition of fire threat index. The unsupervised method of fire region segmentation with Q-rough set has been proven to be effective visually and quantitatively in classifying fire pixels correctly. The performance of Q-rough set based fire segmentation has been proven to be superior in identifying the correct fire region over different types of video sequences with respect to different state of the art methods. The fire threat index is proven to be effective in reflecting the spread of the fire quickly. This index works well with all the fire segmentation methods, but it reflects the threat the best with Q-rough set method. 

Therefore it can be concluded that the proposed methods can be integrated in video surveillance systems to generate quick fire alarm. The Q-rough set based segmentation method could also be used in the other areas of video analysis, like, object tracking since the Q-agents could learn and explore the object of interest by themselves alongwith the input stream data.   Besides, the lower-upper approximations with Q-rough set, defined here, could be used other areas to deal with the incompleteness of knowledge base and stream data.  

The fire threat index is defined here assuming that the surveillance videos will be captured by static cameras. But the method may fail if the surveillance is carried out with moving cameras or ego-centric videos. The definition of the index could be modified in future to make it more general and applicable to any kind of surveillance systems. 

\bibliographystyle{unsrt}
\bibliography{fire}

\begin{thebibliography}{10}

\bibitem{PAWLAK_RS_BOOK_1992}
Z.~Pawlak.
\newblock {\em Rough Sets: Theoretical Aspects of Reasoning about Data}.
\newblock Kluwer Academic Publishers, Norwell, MA, 1992.

\bibitem{Pal_17}
S.~K. Pal and D.~B. Chakraborty.
\newblock Granular flow graph, adaptive rough rule generation and tracking.
\newblock {\em IEEE Trans. on Cyberns.}, 47(12):4096 -- 4107, 2017.

\bibitem{Liu_18}
J.~Liu, Y.~Lin, Y.~Li, W.~Weng, and S.~Wu.
\newblock Online multi-label streaming feature selection based on neighborhood
  rough set.
\newblock {\em Pattern Recognition, Elsevier.}, 84(12):273--287, 2018.

\bibitem{Zhaoa_20}
J.~Zhaoa, J.~Liang, Z.~Dong, D.~Tang, and Z.~Liu.
\newblock Accelerating information entropy-based feature selection using rough
  set theory with classified nested equivalence classes.
\newblock {\em Pattern Recognition, Elsevier}, 107(11):273--287, 2020.

\bibitem{Hasslet_10}
H.~V. Hasselt.
\newblock Double q-learning.
\newblock In J.~D. Lafferty, C.~K.~I. Williams, J.~Shawe-Taylor, R.~S. Zemel,
  and A.~Culotta, editors, {\em Advances in Neural Information Processing
  Systems 23}, pages 2613--2621. Curran Associates, Inc., 2010.

\bibitem{Sutton18}
R.~S. Sutton and A.~G. Barto.
\newblock {\em Reinforcement Learning: An Introduction}.
\newblock The MIT Press, second edition, 2018.

\bibitem{russel2020}
S.~Russell and P.~Norvig.
\newblock {\em Artificial Intelligence: A Modern Approach}.
\newblock Prentice Hall, 4 edition, 2020.

\bibitem{Zadeh_97}
Lotfi~A. Zadeh.
\newblock Toward a theory of fuzzy information granulation and its centrality
  in human reasoning and fuzzy logic.
\newblock {\em Fuzzy Sets Syst.}, 90(2):111--127, 1997.

\bibitem{Gaur_20}
A.~Gaur, A.~Singh, A.~Kumar, A.~Kumar, and K.~Kapoor.
\newblock Video flame and smoke based fire detection algorithms: A literature
  review.
\newblock {\em Fire Technology, Springer}, 56(5):1943--1980, 2020.

\bibitem{Chen_04}
P.~Wu T.~Chen and Y.~Chiou.
\newblock An early fire-detection method based on image processing.
\newblock In {\em Proc. IEEE Int. Image Process.}, pages 1707--1710, 2004.

\bibitem{Fernandesa_04}
A.~M. Fernandesa, A.~B.Utkin, A.~V.Lavrov, and R.~M.Vilar.
\newblock Development of neural network committee machines for automatic forest
  fire detection using lidar.
\newblock {\em Pattern Recognition, Elsevier.}, 37(10):2039--2047, 2004.

\bibitem{Toreyin_05}
Y.~Dedeoglu B.U.~Toreyin and A.E. Cetin.
\newblock Flame detection in video using hidden markov models.
\newblock In {\em Proc. IEEE Int. Conf. Image Process.}, pages 1230--1233,
  2005.

\bibitem{Toreyin_06}
Y.~Dedeoglu B.U.~Toreyin and A.E. Cetin.
\newblock Computer vision based method for real-time fire and flame detection.
\newblock {\em Pattern Recognition Lett.}, 27:49--58, 2006.

\bibitem{Celik_10}
T.~Celik.
\newblock Fast and efficient method for fire detection using image processing.
\newblock {\em ETRI Journal}, 32:881--890, 2010.

\bibitem{Zhao_11}
J.~Zhao, Z.~Zhang, S.~Han, C.~Qu, Z.~Yuan, and D.~Zhang.
\newblock Svm based forest fire detection using static and dynamic features.
\newblock {\em Computer Science and Information Systems}, 8:821–841, 2011.

\bibitem{Chino-15}
D.~Y.~T. {Chino}, L.~P.~S. {Avalhais}, J.~F. {Rodrigues}, and A.~J.~M.
  {Traina}.
\newblock Bowfire: Detection of fire in still images by integrating pixel color
  and texture analysis.
\newblock In {\em SIBGRAPI Conference on Graphics, Patterns and Images}, pages
  95--102, 2015.

\bibitem{Dimitropoulos_15}
K.~Dimitropoulos, P.~Barmpoutis, and N.~Grammalidis.
\newblock Spatio-temporal flame modeling and dynamic texture analysis for
  automatic video-based fire detection.
\newblock {\em IEEE Trans. on Cir. Sys. Vid. Tech.}, 25(2):339--351, 2015.

\bibitem{Li_20}
P.~Li and W.~Zhao.
\newblock Image fire detection algorithms based on convolutional neural
  networks.
\newblock {\em Case Studies in Thermal Engineering, June 2020 , Elsevier},
  19:1--11, 2020.

\bibitem{Muhammad_18}
K.~Muhammad, J.~Ahmad, and S.~W. Baik.
\newblock Early fire detection using convolutional neural networks during
  surveillance for effective disaster management.
\newblock {\em Neurocomputing, Elsevier}, 288:30--42, 2018.

\bibitem{Kim_19}
B.~Kim and J.~Lee.
\newblock A video-based fire detection using deep learning models.
\newblock {\em Appl. Sci., MDPI}, 9:2862 -- 2874, 2019.

\bibitem{Cai_19}
Y.~Cai, Y.~Guo, Y.~Li, H.~Li, and J.~Liu.
\newblock Fire detection method based on improved deep convolution neural
  network.
\newblock In {\em ACM ICCPR: Int. Conf. on Computing and Pattern Recognition},
  pages 466--470, 2019.

\bibitem{Reinhard_05}
E.~Reinhard, G.~Ward, S.~Pattanaik, and P.~Debevec.
\newblock {\em High Dynamic Range Imaging: Acquisition, Display, and
  Image-Based Lighting (The Morgan Kaufmann Series in Computer Graphics)}.
\newblock Morgan Kaufmann Publishers Inc., San Francisco, CA, USA, 2005.

\bibitem{Grammalidis_17}
N.~Grammalidis, K.~Dimitropoulos, and E.~Cetin.
\newblock Firesense database of videos for flame and smoke detection [data
  set].
\newblock In {\em Zenodo}, pages 466--470, 2017.

\bibitem{IE_18}
Inside Edition.
\newblock Daring brides light wedding dresses on fire while wearing them.
\newblock In {\em YouTube}, Link: https://www.youtube.com/watch?v=yNdaJblEqiU,
  2018.

\bibitem{NFA_15}
National Fire~Protection Association.
\newblock Christmas tree fires can turn devastating and deadly within seconds.
\newblock In {\em YouTube}, Link:
  https://www.youtube.com/watch?v=xr6b9b8FYKkt=5s, 2015.

\end{thebibliography}

\end{document}